\documentclass[letterpaper, 10 pt, journal]{ieeeconf}

\IEEEoverridecommandlockouts

\usepackage{etextools}
\usepackage{amssymb}
\usepackage{float}
\usepackage{makeidx}
\usepackage{amsmath}
\usepackage{bbm}
\usepackage{epsfig}
\usepackage{epsf}
\usepackage{psfrag}
\usepackage{verbatim}
\usepackage{color}
\usepackage{multirow}
\usepackage{tabularx}
\usepackage{booktabs}
\usepackage[tight,footnotesize]{subfigure}
\usepackage{array}
\usepackage{soul}
\usepackage{footnote}
\usepackage{cite}
\usepackage{dblfloatfix}
\usepackage{tcolorbox}

\usepackage{color, colortbl}
\usepackage[colorlinks,bookmarksnumbered,citecolor=orange,urlcolor=orange]{hyperref}
\usepackage{graphicx}
\graphicspath{{Figures}}
\DeclareGraphicsExtensions{.pdf,.png}
\usepackage{bigstrut}
\usepackage[english]{babel}

\usepackage{csquotes}

\usepackage[T1]{fontenc}
\usepackage{algorithm, setspace}
\usepackage{algpseudocode}
\usepackage{url}
\usepackage{multirow}
\usepackage{float}
\usepackage{xcolor}
\usepackage{hyperref}
 \hypersetup{
     colorlinks=true,
     linkcolor=orange,
     filecolor=orange,
     citecolor=orange,      
     urlcolor=orange,
     }

\newcommand{\p}[1]{\smallskip \noindent \textbf{{#1}.}}
\newcommand{\eq}[1]{Equation~(\ref{eq:#1})}
\newcommand{\fig}[1]{Figure~\ref{fig:#1}}
\usepackage{balance}
\usepackage{amssymb}
\let\labelindent\relax
\usepackage{enumitem}

\title{\LARGE
Stable-BC: Controlling Covariate Shift with Stable Behavior Cloning
}
\author{Shaunak A. Mehta$^{1}$, Yusuf Umut Ciftci$^{2}$, Balamurugan Ramachandran$^{1}$, Somil Bansal$^{2}$, and Dylan P. Losey$^{1}$
\thanks{This work was supported by NSF Grants \#2246446 and \#2240163 \newline $^{1}$S. A. Mehta, B. Ramachandran, and D. P. Losey are with the Collaborative Robotics Lab (\href{https://collab.me.vt.edu/}{Collab}), Dept. of Mechanical Engineering, Virginia Tech, Blacksburg, VA 24061. \newline $^{2}$Y. U. Ciftci and S. Bansal are with the Safe and Intelligent Autonomy Lab (\href{https://smlbansal.github.io/sia-lab/index.html}{SIA Lab}), USC Viterbi School of Engineering, Los Angeles, CA 90089. \newline Email: \texttt{mehtashaunak@vt.edu, yciftci@usc.edu}}
}

\begin{document}

\maketitle

\begin{abstract}

Behavior cloning is a common imitation learning paradigm.
Under behavior cloning the robot collects expert demonstrations, and then trains a policy to match the actions taken by the expert.
This works well when the robot learner visits states where the expert has already demonstrated the correct action; but inevitably the robot will also encounter new states outside of its training dataset.
If the robot learner takes the wrong action at these new states it could move farther from the training data, which in turn leads to increasingly incorrect actions and compounding errors.
Existing works try to address this fundamental challenge by augmenting or enhancing the training data.
By contrast, in our paper we develop the \textit{control theoretic} properties of behavior cloned policies.
Specifically, we consider the error dynamics between the system's current state and the states in the expert dataset.
From the error dynamics we derive model-based and model-free conditions for stability: under these conditions the robot shapes its policy so that its current behavior converges towards example behaviors in the expert dataset.
In practice, this results in Stable-BC, an easy to implement extension of standard behavior cloning that is provably robust to covariate shift.
We demonstrate the effectiveness of our algorithm in simulations with interactive, nonlinear, and visual environments.
We also conduct experiments where a robot arm uses Stable-BC to play air hockey.
See our website here: \url{https://collab.me.vt.edu/Stable-BC/} 

\end{abstract}

\vspace{-0.9em}
\section{Introduction}

\begin{figure}[t]
	\begin{center}
 		\includegraphics[width=0.9\columnwidth]{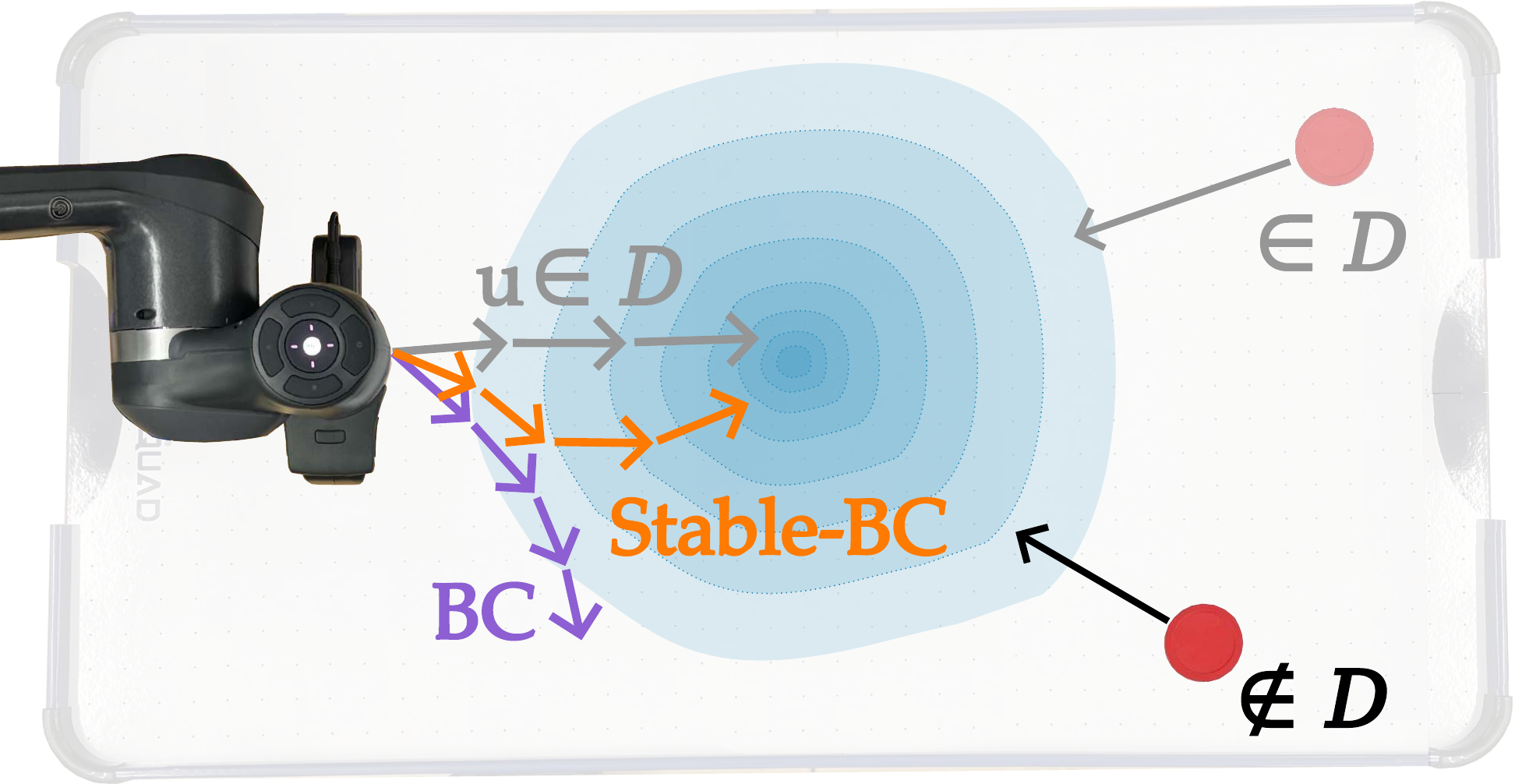}
		\caption{Robot playing air hockey by behavior cloning demonstrations $\mathcal{D}$. The robot can successfully hit the puck when it moves at an angle and velocity observed during training ($\in \mathcal{D}$). However, when the puck moves at new angles or velocities ($\notin \mathcal{D}$), standard behavior cloning (BC) misses the puck entirely because of covariate shift. To address this problem we introduce Stable-BC, a variant of BC that encourages the system state to evolve similarly to the expert's demonstrated behaviors.}
        \vspace{-2.0em}
		\label{fig:front}
	\end{center}
\end{figure}

Behavior cloning enables robots to learn new tasks by imitating humans.
Consider the air hockey game in \fig{front}.
Here a human might show the robot arm a few examples of how to block the puck.
With behavior cloning, the robot learns to match the human's actions, so that --- if the robot sees the puck moving like it did in one of the examples --- the robot mimics how the human expert blocked that puck.
But what happens when the puck moves in a new way (e.g., with a previously unseen angle or velocity)?
This is an instance of \textit{covariate shift}, a difference between what the robot observes at training time and what the robot encounters when executing its learned policy \cite{ross2010efficient, de2019causal, spencer2021feedback}.
Covariate shift is a fundamental problem for behavior cloning because it can lead to \textit{compounding errors}: a small change in the puck angle or velocity may cause the robot to take the wrong action, resulting in a larger covariate shift and increasingly incorrect robot behavior (i.e., the robot misses the puck entirely).

Existing research tries to prevent compounding errors and learn robust policies by focusing on the \textit{data} (i.e., the human examples) used during behavior cloning.
For instance, when the human provides initial demonstrations of their desired behavior, off-policy methods perturb the human to collect more diverse examples \cite{laskey2017dart, umut2024safe}, or synthetically augment the human's demonstrations to gather a larger dataset \cite{ke2023ccil, ke2021grasping, zhou2023nerf, chang2021mitigating, hoque2024intervengen}.
Similarly, as the robot executes what it has learned in the environment, on-policy methods encourage the human to correct the robot when it makes mistakes, and then add these new examples to the robot's dataset \cite{ross2011reduction, hoque2021thriftydagger, kelly2019hg, menda2019ensembledagger, spencer2022expert, mehta2023unified}.
In either case, a core idea within existing works is that --- if the robot has sufficient and relevant data --- it will return to the desired behavior when it encounters novel states and situations.

In this paper we introduce an alternate viewpoint for robust behavior cloning.
Instead of enhancing the data used to train the robot, we will focus on the control theoretic properties of the robot's learned policy.
We hypothesize that:
\begin{center}\vspace{-0.3em}
\textit{Behavior cloned policies are robust when they stick to \\what they know, i.e., when the robot remains close to the example motions that the expert demonstrated}.
\vspace{-0.3em}
\end{center}
Our intuition here is that compounding errors occur when the robot drifts away from its training distribution, and so we can avoid these errors by intentionally attracting the robot back towards the dataset.
We apply this hypothesis to formulate covariate shift as a dynamical system, and derive stability conditions that cause the robot to converge towards behaviors the human has demonstrated.
In practice, this results in a behavior cloning algorithm with two loss functions.
First, the standard loss that causes the robot policy to match the human's behavior, and second, a stability loss that makes the expert dataset a basin of attraction.
Robots that execute this learned policy are inherently robust to covariate shift: e.g., when the puck's angle and velocity in \fig{front} deviates from the dataset, the robot extrapolates from the expert examples while still remaining similar to these demonstrated behaviors.

Overall, we make the following contributions:

\p{Stability Condition} 
We write covariate shift as a linearized dynamical system.
We then use these error dynamics to derive model-based and model-free stability conditions.
These conditions bound the covariate shift in the robot state as a function of the covariate shift in the environment state.

\p{Stable-BC} 
We leverage our analysis to develop Stable-BC, a behavior cloning algorithm that encourages policy stability around the dataset.
Two key advantages of our approach are (a) it is easy to implement and (b) it does not alter the training data.
As such, Stable-BC can be applied independently or alongside existing on-policy or off-policy methods.

\p{Experiments}
We conduct simulated experiments in interactive, nonlinear, and visual environments. We also perform a real-world study where human users teach a robot arm to play a simplified game of air hockey. Across all experiments, we find that Stable-BC results in more robust and effective policies than state-of-the-art baselines.

\section{Related Works} \label{sec:related}

Behavior cloning is a common imitation learning approach in robotics \cite{pomerleau1988alvinn}. 
Below we summarize recent methods for robust behavior cloning, as well as works that apply stability analysis in similar imitation learning settings. 

\p{Off-Policy Approaches}
Within behavior cloning the robot is given a dataset of expert state-action pairs, and the robot learns a policy to match this dataset. 
One way to enhance the robustness of the learned policy is to improve the quality of the original dataset. 
For example, some off-policy methods intentionally perturb the human when they are providing demonstrations, and then record how the human corrects the robot in response to those disturbances \cite{laskey2017dart, umut2024safe}.
Other approaches synthetically augment the dataset provided by the human to enhance the diversity and quantity of training data \cite{ke2023ccil, ke2021grasping, zhou2023nerf, chang2021mitigating, hoque2024intervengen}. 
For example, Ke \textit{et al.} \cite{ke2023ccil} learn a dynamics model from the data, and then use that model to generate new state-action pairs that align with the expert's demonstrations.

\p{On-Policy Approaches}
The robot can also iteratively gather new data by executing its policy in the environment, and then asking the human to provide expert guidance when the robot makes mistakes \cite{ross2011reduction}.
Recent works in interactive imitation learning explore when to query the human for additional data, and how to best leverage that data to accelerate the robot's learning \cite{kelly2019hg, menda2019ensembledagger, hoque2021thriftydagger, mehta2023unified, spencer2022expert}. 
For example, in Menda \textit{et al.} \cite{menda2019ensembledagger} the robot maintains an ensemble of behavior cloned models; the robot asks for human guidance when it encounters a state where the models disagree over which action to take.

Overall, both off-policy and on-policy methods focus on the data used to train the behavior cloned agent.
This is orthogonal to our approach, where we will only consider the control theoretic properties of the learned policy. 

\p{Stable Imitation Learning}
Multiple related works have applied stability analysis to imitation learning. 
For example, safety filters \cite{wabersich2023data, hsu2023safety, reichlin2022back} monitor the robot's real-time error (e.g., covariate shift), and revert to a given or learned backup policy when that error exceeds some threshold.
Other methods structure the robot's policy like a dynamical system, and ensure that this system converges towards a goal state \cite{totsila2023end, khansari2011learning, mehta2024strol}.
The robot can also take advantage of the stability of the human teacher: Pfrommer \textit{et al.} \cite{pfrommer2022tasil} train the robot's behavior cloned policy to match the higher order derivatives of the human's demonstrations.

Our approach is most similar to recent research by Kang \textit{et al.} \cite{kang2022lyapunov}.
In \cite{kang2022lyapunov} the authors seek to prevent distribution shift in learned policies by formulating the entire training distribution as an equilibrium point.
However, \cite{kang2022lyapunov} focuses on model-based reinforcement learning --- by contrast, we study local stability for behavior cloning.
\section{Problem Statement} \label{sec:problem}

We consider settings where a robot is learning to imitate human behaviors.
This includes scenarios where the robot is acting in isolation (e.g., a robot arm reaching a goal position), as well as interactive tasks where the robot must reason over other agents (e.g., an autonomous vehicle at an intersection with a human-driven car).
Offline the robot is provided with a dataset of state-action pairs, and the robot applies behavior cloning to learn its policy.
Our objective is for the robot to extrapolate from the dataset to the learned policy, so that online --- when the robot executes this policy --- the system is robust to covariate shift.

\p{Robot} Let $x \in \mathcal{X} \subset \mathbb{R}^{m}$ be the state of the robot and let $u \in \mathcal{U} \subset \mathbb{R}^{n}$ be the robot's action.
For instance, within our air hockey example from \fig{front}, $x$ is the position of the paddle and $u$ is the robot's end-effector velocity.
The robot's state evolves according to the dynamics $\dot x(t) = f\big(x(t), u(t)\big)$.
We assume that the robot can observe its state $x$, and that the robot has an accurate model of its own dynamics $f$.

\p{Environment} During each task the robot interacts with its environment.
This environment could consist of static goals, dynamic objects, or even other agents.
We separate the state of the environment from the state of the robot: let $y \in \mathcal{Y} \subset \mathbb{R}^{d}$ be the environment's state.
Returning to our air hockey example, $y$ could be the angle and velocity of the puck the robot is trying to block.
The environment state $y$ updates according to its dynamics $\dot y(t) = g\big(x(t), y(t), u(t)\big)$.
We assume that the robot can observe the environment state $y$, but we \textit{do not assume} that the robot has access to the environment dynamics $g$.
For instance, when the robot arm moves to block the puck, the robot does not know how the angle and velocity of the puck might change after collision. 

\p{Behavior Cloning} The robot is given a dataset $\mathcal{D}$ of $N$ state-action pairs. 
Here the overall system state $(x, y)$ consists of both the robot's state and the environment's state. 
Hence, the offline dataset is: $\mathcal{D} = \{(x_1, y_1, u_1), \ldots, (x_N, y_N, u_N)\}$.

Based on this dataset, the robot should learn a policy $\pi$ that maps from system states to robot actions: $\pi(x, y) \rightarrow u$.
We instantiate this policy as a neural network with weights $\theta$.
Within standard behavior cloning algorithms the robot learns $\theta$ such that the policy's actions match the human's actions across states in the dataset.
More specifically, the robot learns $\theta$ to minimize the loss function:
\begin{equation} \label{eq:P1}
    \mathcal{L}_{BC}(\theta) =  \sum_{(x, y, u) \in \mathcal{D}} \|\pi_\theta(x, y) - u\|^2
\end{equation}
Although the resulting policy is designed to mimic the human at states within dataset $\mathcal{D}$, it may fail to match the human outside of this dataset.
Consider \fig{front}: training the robot using \eq{P1} leads to compounding errors when the puck moves at a previously unseen angle and velocity.

\section{Stable Behavior Cloning} \label{sec:method}

Our objective is to learn a robot policy that maintains the correct behavior despite covariate shift.
Here we return to our hypothesis: \textit{behavior cloning is robust when robots stick to what they know}, i.e., when robots remain close to the example motions the expert has demonstrated.
We will apply this hypothesis to introduce \textit{Stable-BC}, a control theoretic approach for behavior cloning. 
Stable-BC is based on the error dynamics between the system's current state and the states in dataset $\mathcal{D}$.
We define these error dynamics in Section~\ref{sec:method1}, and derive the stability conditions under which the error locally converges to zero.
In practice, robot policies that satisfy these stability conditions update the robot's state to converge towards behaviors in the dataset, mitigating the effects of covariate shift and preventing compounding errors.
We next apply our stability analysis to derive new loss functions for behavior cloning when the robot has access to a model of the environment (Section~\ref{sec:method2}), and when the environment dynamics are unknown (Section~\ref{sec:method3}).

\subsection{Error Dynamics and Stability Analysis} \label{sec:method1}

We start by formulating the error dynamics between the current state of the system and the states observed during training.
Let $(x', y')$ be the current state, and let $(x, y) \in \mathcal{D}$ be a labeled state from the training dataset.
In order to remain close to behaviors that the human expert has demonstrated, the robot should take actions so that a trajectory starting at $(x', y')$ converges towards a trajectory starting at $(x, y)$.
Recall that the robot state $x$ updates according to the dynamics $\dot x = f(x, u)$, and the environment state $y$ updates according to dynamics $\dot y = g(x, y, u)$.
Utilizing these equations, the overall error dynamics become:
\begin{align} \label{eq:Mx}
    \dot x' - \dot x &= f\big(x', u'\big) - f\big(x, u\big) \\
    \label{eq:My}
    \dot y' - \dot y &= g\big(x', y', u'\big) - g\big(x, y, u\big)
\end{align}
Compounding errors occur when $(x', y')$ diverges from the dataset $(x, y) \in \mathcal{D}$, i.e., as the error grows over time.
To mitigate compounding errors we therefore want to design the robot's policy such that Equations~(\ref{eq:Mx}) and (\ref{eq:My}) converge to zero.
Substituting $u = \pi(x, y)$ into the above equations, where $\pi$ is the robot's policy, and applying a first order Taylor Series approximation around $(x, y) \in \mathcal{D}$, we reach:
\begin{align*}
    \dot x' - \dot x = \big(\nabla_x f + \nabla_u f \cdot \nabla_x \pi\big)(x' - x)\\
    + \big(\nabla_u f \cdot \nabla_y \pi\big)(y' - y)\\
    \dot y' - \dot y = \big(\nabla_x g + \nabla_u g \cdot \nabla_x \pi\big)(x' - x)\\
    + \big(\nabla_y g + \nabla_u g \cdot \nabla_y \pi\big)(y' - y)
\end{align*}
It is important to recognize that these equations are \textit{coupled}. 
The covariate shift in the robot state $x$ is a function of the covariate shift in the environment state $y$, and vice versa. 
This coupling is reflected in our motivating example from \fig{front}: if the puck moves with a new angle or velocity such that $y'$ is dissimilar from any $y$ in the dataset (i.e., $\|y' - y\|$ is large), then this could cause the robot's position $x'$ to increasingly diverge from labeled states $x$.

Below we write the coupled system in standardized form:
\begin{align} \label{eq:m2}
    \dot z = Az 
\end{align}
where $z$ is the augmented error state, and $A$ is a square matrix that captures the coupled error dynamics:
\begin{align} \label{eq:m3}
    z &= \begin{bmatrix} x' - x \\ y' - y \end{bmatrix},\\
    \label{eq:ma}
    A &= \begin{bmatrix} \nabla_x f + \nabla_u f \cdot \nabla_x \pi & \nabla_u f \cdot \nabla_y \pi \\ \nabla_x g + \nabla_u g \cdot \nabla_x \pi & \nabla_y g + \nabla_u g \cdot \nabla_y \pi \end{bmatrix}
\end{align}
Ideally, we want the error dynamics $\dot z = Az$ to be \textit{stable} about the equilibrium $z = 0$.
If we achieve this stability, then the the behavior starting at the current system state $(x', y')$ will converge towards the demonstrated behavior starting at a labeled state $(x, y)$.
This mathematically formalizes our original hypothesis: stabilizing \eq{m2} encourages the robot to take actions that remain close to the examples the expert has demonstrated.
Because matrix $A$ is a local, linearized approximation of the error dynamics, we conclude that $\dot z = Az$ is \textit{locally} stable if and only if matrix $A$ is stable, i.e., if all eigenvalues of $A$ have negative real parts \cite{teschl2012ordinary} (Theorem $7.1$).
When $A$ is stable the system locally converges towards $z=0$.
The rate of convergence is determined by the eigenvalues of matrix $A$, where more negative eigenvalues result in faster convergence \cite{teschl2012ordinary} (Theorem $7.2$).

\subsection{Stable-BC for Model-Based Settings} \label{sec:method2}

Overall, our analysis from Section~\ref{sec:method1} indicates that we can mitigate errors between the system's current behavior and the expert's demonstrated behaviors by ensuring that matrix $A$ is stable.
Inspecting \eq{ma}, we find that $A$ depends upon the robot dynamics $f$, the robot policy $\pi$, and the environment dynamics $g$.
In this section we will focus on \textit{model-based settings} where the robot has access to all of these terms.
Put another way, here we assume that the robot not only knows its own dynamics $f(x,u)$, but it also has an accurate model of the environment dynamics $g(x, y, u)$.

\p{Stability vs. Performance} Within model-based settings the robot can directly compute the $A$ matrix.
Accordingly --- in order to make the matrix $A$ locally stable --- we simply need to train the robot's policy $\pi$ such that all the eigenvalues of $A$ have negative real parts across each state $(x,y) \in \mathcal{D}$.
But just ensuring that $A$ is stable does not mean that the robot has learned to perform the task correctly.
In fact, this stability can conflict with performance; for instance, when $A$ is stable the robot may converge to $z=0$, and then remain at rest at that local equilibrium instead of continuing to complete the task.
In practice, we resolve this theoretical conflict between stability and performance by training the robot to match the expert's demonstrations (standard behavior cloning loss) while also penalizing the robot's policy when $A$ is unstable (our proposed addition).
This leads to the loss function:
\begin{align} \label{eq:m4}
    \begin{split}
        \mathcal L(\theta) = \sum_{(x, y, u) \in \mathcal D} \Bigg[&\|u -\pi_\theta(x, y)\|^2 \\
        &+ \lambda \sum_{\sigma_i \in eig(A)} ReLU(Re(\sigma_i))\Bigg]
    \end{split}
\end{align}
The first term in \eq{m4} matches the original behavior cloning loss function from \eq{P1}.
Within the second term, $\sigma_i \forall i\in \{1,2, \cdots\}$ are the eigenvalues of $A$, $Re(\sigma)$ represents the real part of eigenvalue $\sigma$, and $ReLU$ is the Rectified Linear Unit activation. 
The constant $\lambda > 0$ is a hyperparameter set by the designer that determines the relative weight of the two loss terms.
Intuitively, a robot policy that minimizes \eq{m4} mimics the expert's actions across the dataset $\mathcal{D}$, while also shaping its policy to converge towards the demonstrated behaviors.
We directly use this loss function to train Stable-BC policies in model-based settings where the robot has an estimate of $g$.

\vspace{-0.5em}
\subsection{Stable-BC for Model-Free Settings} \label{sec:method3}

In practice, often robots do not have a model of how their actions will impact the environment around them.
For instance, in our air hockey experiments the robot arm does not know how its position $x$ and velocity $u$ will change $y$, the angle and velocity of the puck.
In this section we therefore consider \textit{model-free settings} where the robot is not given the environment dynamics $g(x, y, u)$.
Model-free settings are challenging because --- without a model of $g$ --- the robot can only compute the top row of the $A$ matrix in \eq{ma}.
In general, if we have no information about the environment dynamics $g$ or state $y'$, we cannot make guarantees about the overall stability of matrix $A$. 

\p{Bounded Stability}
To resolve this issue we define $\|y' - y\|$ as the magnitude of the environment's covariate shift.
In many settings it is reasonable to assume that this magnitude is \textit{bounded}, i.e., the environment in which the robot is performing its task is similar to the environments seen during training.
Moving forward, we will therefore assume that the magnitude of the environment's covariate shift has some upper bound $\|y' - y\| \leq \epsilon$.
Let $A_1 = \nabla_x f + \nabla_u f \cdot \nabla_u \pi$ and $A_2 = \nabla_u f \cdot \nabla_y \pi$. 
Substituting these terms in \eq{ma}, the first row of the $A$ matrix can be written as:
\begin{equation}
    \dot e_x = A_1 e_x + A_2 e_y, \quad e_x = x' - x, \quad e_y = y' - y
\end{equation}
Integrating both sides of this ordinary differential equation, we obtain: $e_x(t) = e_x(0) e^{A_1t} + \int_{\tau=0}^t A_2 e_y(\tau) e^{A_1(t-\tau)} d\tau$.
Taking the matrix norm of the result, and substituting in the upper bound for $\|e_y\| \leq \epsilon$, we obtain an upper bound on the covariate shift in the robot's state:
\begin{equation} \label{eq:mz}
    \|e_x(t)\| \leq \|e_x(0)e^{A_1t}\| + \|A_2\| \epsilon \int_{\tau = 0}^t \|e^{A_1(t - \tau)}\| d\tau    
\end{equation}
Intuitively, \eq{mz} provides an limit on how far the robot state at test time, $x'$, could diverge from the robot states during training, $x$.
Similar to our approach from Section~\ref{sec:method1}, our goal here is to minimize the upper bound on this error and cause $\|e_x\|$ to converge towards zero.

In order to minimize the right side of \eq{mz} we propose to design the robot's policy $\pi$ such that matrix $A_1$ is stable.
This causes $\|e_x(0)e^{A_1t}\| \to 0$, and thus \eq{mz} simplifies to $\|e_x(t)\| \leq \|A_2\| \epsilon \int_{\tau = 0}^t \|e^{A_1(t - \tau)}\| d\tau$.
Next, leveraging the properties of matrix exponentials, we have that $\|e^{ A_1 t}\| \leq e^{\|A_1\|t}$, where $\|A_1\|$ is the induced $2$-norm of matrix $A_1$. Substituting this inequality back into the equation and solving the integral, we finally reach:
\begin{align} \label{eq:m7}
    \|e_x(t)\| \leq \dfrac{\|A_2\| \cdot \epsilon}{\|A_1\|} \cdot \Big(e^{\|A_1\|t} - 1\Big)
\end{align}
\eq{m7} offers a useful upper bound on the robot's state error.
Provided that the change in the environment state is bounded by $\epsilon$, \eq{m7} shows that the covariate shift in the robot's state is also bounded, and the magnitude of that bound depends on the off-diagonal matrix $A_2$.
For instance, if we design the matrix $A$ such that $\|A_2\| \to 0$, then the upper bound on $\|e_x(t)\|$ also converges towards zero, and the robot's behavior at test time will remain similar to the examples given at training time.

\begin{algorithm}[t]
    \caption{Stable-BC}
    \label{alg:M1}
    \begin{algorithmic}[1] 
        \State Given: state-action pairs $\mathcal{D}$ and robot dynamics $f$
        \State Initialize: robot policy $\pi_\theta(x, y)$ with weights $\theta$
        \For {$i \in 1, 2, \cdots$}
            \If {$g$ is available}
               \State Compute loss $\mathcal{L(\theta)}$ using \eq{m4}
            \Else
                \State Compute loss $\mathcal{L(\theta)}$ using \eq{m8}
            \EndIf
            \State Update robot policy $\theta \gets \theta - \alpha \nabla_\theta \mathcal{L}(\theta)$            
        \EndFor
        \State \Return Trained robot policy $\pi_\theta(x,y)$
    \end{algorithmic}
\end{algorithm}

\p{Stability vs. Performance} To summarize our analysis, in model-free settings we cannot directly stabilize matrix $A$. Instead, we enforce an upper bound on the covariate shift in the robot's state by designing policy $\pi$ such that:
\begin{enumerate}
    \item All eigenvalues of matrix $A_1$ (the top left component of $A$) have negative real parts
    \item The magnitude of matrix $A_2$ (the top right component of $A$) is minimized
\end{enumerate}
We note that the robot can compute both $A_1$ and $A_2$, since neither term depends on the environment dynamics $g$.

Examining these two conditions we again find a conflict between stability and performance.
Specifically, if $\|A_2\| \to 0$, then $\nabla_y \pi \to 0$ and the robot's policy no longer depends upon the environment state $y$.
From a stability perspective, this is desirable because $\|A_2\| = 0$ means that the equations in \eq{m2} are decoupled, and thus any error $y' - y$ will not impact $x' - x$.
From a performance perspective, however, this is undesirable because we often need the robot to make decisions based on its environment state.
Consider our air hockey example: to successfully hit the puck, the robot's policy must reason over the state of that puck.
Similar to Section~\ref{sec:method2}, we practically resolve this conflict by training a robot policy that trades-off between mimicking the expert's actions and satisfying the two stability conditions.
Our loss function for Stable-BC in model-free settings is:
\begin{align} \label{eq:m8}
    \begin{split}
        \mathcal L(\theta) = \sum_{(x, y, u) \in \mathcal D} \Bigg[&\|u -\pi_\theta(x, y)\|^2 + \lambda_1 \|A_2\| \\
        &+ \lambda_2 \sum_{\sigma_i \in eig(A_1)} ReLU(Re(\sigma_i))\Bigg]
    \end{split}
\end{align}
where $\sigma$ are the eigenvalues of the sub-matrix $A_1$.
The first term in \eq{m8} matches the original behavior cloning loss function from \eq{P1}. The other terms enforce our two conditions for bounded stability, and $\lambda_1$ and $\lambda_2$ are hyperparameters selected by the designer.
We note that this result for the model-free case is weaker than in the model-based case.
Within Section~\ref{sec:method2} we provided conditions for local asymptotic stability in both $e_x$ and $e_y$; by contrast, here we can only ensure bounded stability for $e_x$ assuming an upper bound on the magnitude of $e_y$.

\p{Algorithm Summary} Given a dataset of state-action pairs $\mathcal{D} = \{(x_1,y_1,u_1) \ldots (x_N, y_N, u_N)\}$ and robot dynamics $f$, the robot uses the procedure outlined in Algorithm~\ref{alg:M1} to learn a behavior cloned policy that is locally stable around the expert demonstrations. 
We refer to this approach as Stable-BC. 
If the robot has access to the environment dynamics $g$, then we use \eq{m4} as the loss function. 
Alternatively, if the environment dynamics are unknown, the robot leverages \eq{m8} as the loss function.
\begin{figure*}[t!]
	\begin{center}
 		\includegraphics[width=1.8\columnwidth]{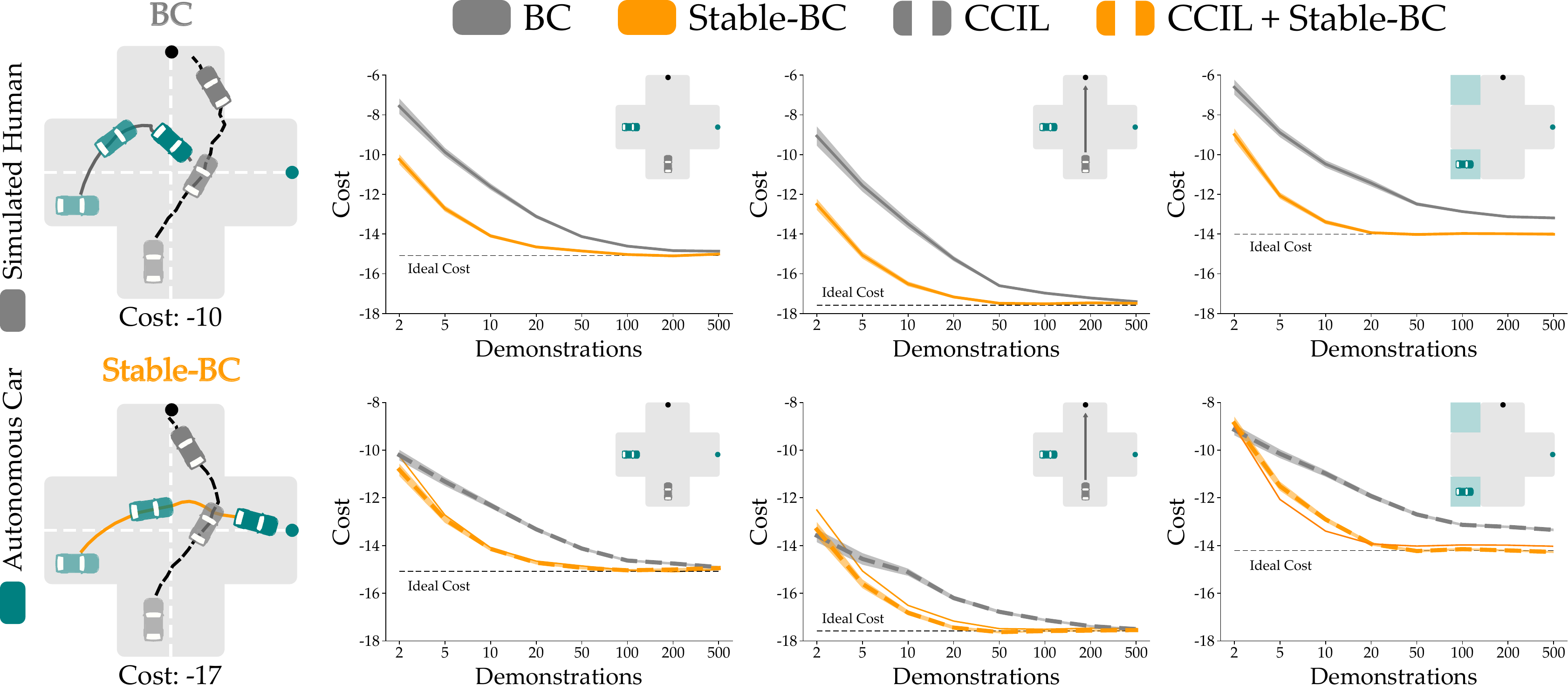}
		\caption{Simulation results from interactive driving. (Left) An example rollout using BC and Stable-BC. With BC the autonomous car gets stuck in the middle of the intersection. By contrast, when using Stable-BC the autonomous car lets the human pass and then crosses afterwards, resulting in a lower cost. (Right) Average cost over $100$ trials as a function of the number of expert demonstrations. In the left column the testing environment matches the training environment. In the middle column the human agent ignores the autonomous car, and in the right column the autonomous car starts from initial states outside of its training distribution. Shaded regions show SEM. Ideal cost is the best-case scenario where the autonomous car's learned policy exactly matches the policy of the human teacher. In the bottom row we plot Stable-BC (solid orange) and CCIL + Stable-BC (dashed orange).}
		\label{fig:sim1}
	\end{center}
    \vspace{-1.5em}
\end{figure*}

\section{Simulations} \label{sec:sims}

In Section~\ref{sec:method} we presented Stable-BC, our method for shaping behavior cloned policies such that they have stable error dynamics.
Next we will perform controlled simulations that compare Stable-BC to standard behavior cloning and recent off-policy variants.
We consider three different tasks: (Section~\ref{sec:sim1}) an interactive driving environment where an autonomous car and human vehicle are trying to cross an intersection, (Section~\ref{sec:sim2}) a single-agent quadrotor environment where drone with nonlinear dynamics must safely navigate $3$D spaces, and (Section~\ref{sec:sim3}) a simple visual setting where a point mass uses RGB images to estimate its goal position.
The code for implementation can be found here: 
\url{https://github.com/VT-Collab/Stable-BC}

\vspace{-1em}
\subsection{Interactive Driving} \label{sec:sim1}

In our first simulation an autonomous car learns to cross an intersection while avoiding a human driver (see \fig{sim1}).
This environment is challenging for the autonomous car because it is \textit{interactive}: even if the autonomous car matches the demonstrated behavior, changes in how the human drives during policy execution can lead to covariate shift.

\p{Environment}
The autonomous car's state $x \in \mathbb{R}^2$ is its position, and action $u \in \mathbb{R}^2$ is the autonomous car's velocity. 
State $x$ updates with the known linear dynamics $\dot{x} = u$.
During each interaction the autonomous car tries to reach a static goal position on the opposite side of the intersection while maintaining a safe distance from a human driver.
Here $y \in \mathbb{R}^2$ is the position of the human's vehicle, and $y$ evolves with unknown and nonlinear dynamics $g(x, y)$.

The autonomous car is given a dataset $\mathcal{D}$ of offline demonstrations.
In each demonstration two simulated humans show how both vehicles should navigate the intersection.
The initial car positions $x$ and $y$ are uniformly randomly sampled from regions on the left and bottom of the intersection.
Then one simulated human expert drives the autonomous car while noisily optimizing the following cost function:
\begin{multline} \label{eq:S1}
    Cost(x,y,c) = \| x(t+\Delta t) - c\| - \| x(t) - c \| + \\ 0.75 \cdot \| x(t) - y(t) \| - 0.75 \cdot \|x(t+\Delta t) - y(t)\|
\end{multline}
where $c$ is the constant goal position.
The first two terms of \eq{S1} encourage the car to move towards goal $c$, and the final two terms penalize actions that get closer to the other vehicle $y$.
Simultaneously, a second simulated human controls the human-driven car while optimizing the same cost function (where $x$ and $y$ are switched).
Each individual demonstration results in $20$ state-action pairs $(x,y,u)$.

\p{Methods}
The autonomous car learns from these demonstrations using four different methods.
We start with standard behavior cloning (\textbf{BC}) trained using \eq{P1}.
Next, we implement \textbf{CCIL} \cite{ke2023ccil}: CCIL is a state-of-the-art off-policy approach that builds a dynamics model from the dataset, and then leverages that model to synthetically increase the number of expert state-action pairs.
We compare these baselines to our approach applied independently (\textbf{Stable-BC}), as well as our approach applied alongside CCIL. 
In \textbf{CCIL + Stable-BC} we first use CCIL to enhance the offline dataset, and then train Stable-BC on this augmented dataset.

\p{Results} Our results are summarized in \fig{sim1}.
We report the autonomous car's total cost across an interaction, where the cost at the current timestep is computed using \eq{S1}.
In the top row we compare BC to Stable-BC, and in the bottom row we compare CCIL and CCIL + Stable-BC. 
We also plot Stable-BC in the bottom row for reference.

To evaluate the robustness of the learned policies, we executed each trained policy in three different testing environments.
We started with a testing environment that exactly matched the training environment (left column).
Next, we modified the dynamics $g$ of the human-driven car (middle column).
Instead of moving to avoid the autonomous car with dynamics $g(x,y)$, now the simulated human was self-centered, and only reasoned over their own state using dynamics $g(y)$.
Finally, we sampled the autonomous car's initial state $x(0)$ from regions outside of the training distribution (right column).
The human again used the training dynamics $g(x,y)$, but the autonomous car had to navigate around that human from new regions of the workspace.

\p{Summary} Across all testing environments, our results indicate that Stable-BC outperforms BC and CCIL, and that the differences between Stable-BC and CCIL + Stable-BC are negligible.
This is a positive result because it suggests that our approach is more robust to covariate shift in interactive settings, and that off-policy data-augmentation methods may not be necessary when applying Stable-BC.

\subsection{Nonlinear Quadrotor Navigation} \label{sec:sim2}

\begin{figure}
    \centering
    \includegraphics[width=1\columnwidth]{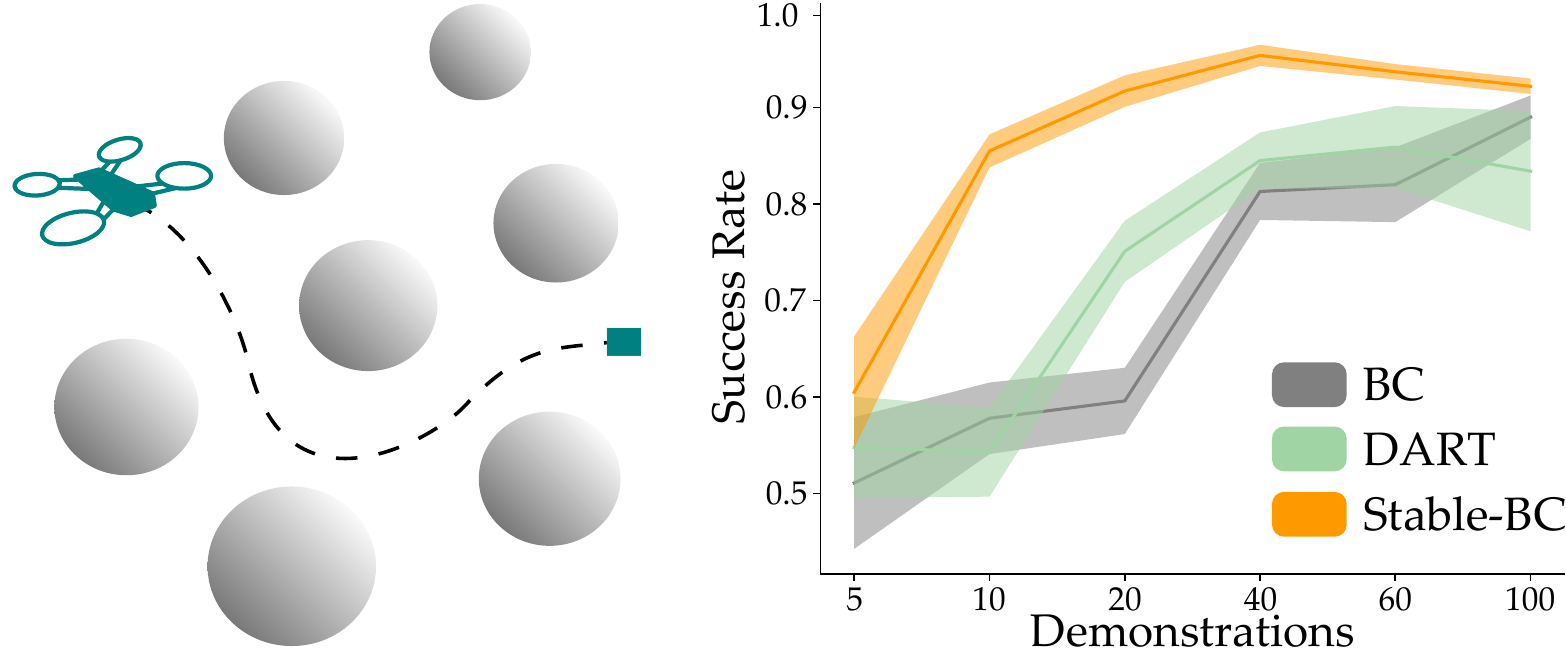}
    \caption{Simulation results for nonlinear quadrotor navigation. (Left) An example trajectory of the quadrotor flying around the $3$D obstacles to reach its goal position. (Right) Average success rate of the quadrotor. We trained the system end-to-end $10$ separate times, and then performed $100$ test rollouts with each trained model. Shaded regions show SEM.}
    \vspace{-1.5em}
    \label{fig:sim2}
\end{figure}

In our second simulation we apply Stable-BC to a nonlinear system.
Specifically, we consider a quadrotor that must navigate across a room with spherical obstacles (see \fig{sim2}).
The quadrotor's state $x \in \mathbb{R}^6$ includes its position $(p_x, p_y, p_z)$ and velocity $(v_x, v_y, v_z)$, and the quadrotor's action $u \in \mathbb{R}^3$ includes its acceleration $u_T$, roll $u_\phi$, and pitch $u_\theta$. 
State $x$ evolves with nonlinear dynamics:
\begin{gather*}
   \dot{p}_x = v_x, \quad \dot{p}_y = v_y, \quad \dot{p}_z = v_z \\
    \dot{v}_x = a_g \tan u_\theta, \quad \dot{v}_y = -a_g \tan u_\phi, \quad \dot{v}_z = u_T - a_g 
\end{gather*}
At the start of each interaction the quadrotor is uniformly randomly initialized on one side of the room. 
There are seven static obstacles that the robot must avoid as it navigates to its fixed goal location on the opposite side of the room. The interaction ends when the quadrotor either reaches within $0.5$ units of the goal (a success) or collides with an obstacle or wall of the room (a failure).

\p{Methods}
Because the obstacles and goal locations are fixed, and the quadrotor knows its dynamics, in this simulation we apply our \textit{model-based approach} for \textbf{Stable-BC}.
We compare Stable-BC to two baselines: \textbf{BC} and \textbf{DART} \cite{laskey2017dart}.
DART is a state-of-the-art data collection approach that perturbs the expert while they provide demonstrations to increase dataset diversity.
As the robot collects expert demonstrations offline, DART iteratively estimates the errors between the expert's actions and its current policy.
DART then injects noise based on these errors when collecting new demonstrations from the expert; this causes the expert to show the robot more diverse and corrective behaviors.
BC and Stable-BC are trained using the same offline dataset that does not include DART's perturbation procedure.
To test the robustness of the learned policies and simulate real-world conditions, we inject Gaussian noise into quadrotor's actions at test time.

\p{Results}
Our results are summarized in \fig{sim2}. 
We report the success rate, i.e., the fraction of trials where the quadrotor reached its goal without collisions.
For all methods the success rate increases when the robot is given more expert demonstrations.
However, Stable-BC achieves a higher success rate with fewer demonstrations as compared to the baselines.
Looking specifically at DART and Stable-BC, we find that Stable-BC with the original offline dataset converges to best-case performance more rapidly than robots which use DART to perturb the expert and collect more diverse data.
These results demonstrate that Stable-BC can be effectively applied to nonlinear systems.

\vspace{-0.5em}
\subsection{Point Mass with Visual Observations} \label{sec:sim3}

\begin{figure}[t!]
	\begin{center}
 		\includegraphics[width=0.95\columnwidth]{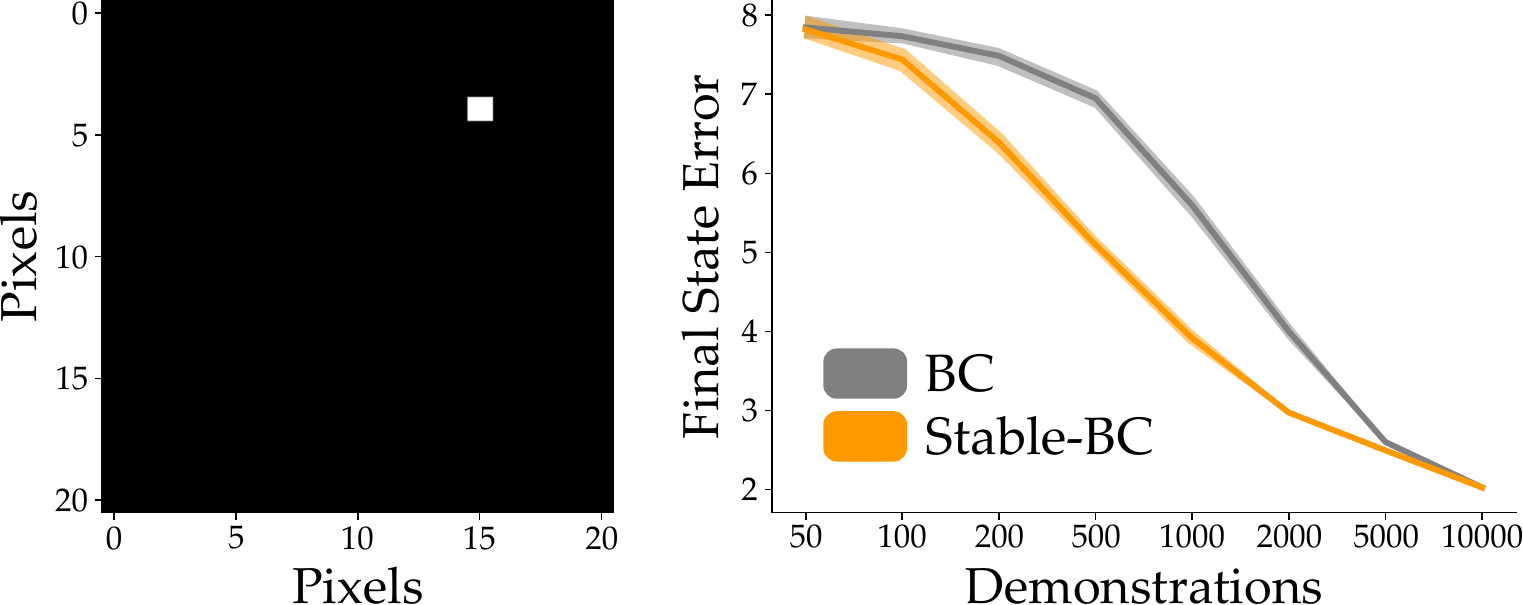}
		\caption{Simulation results for visual observations. (Left) The robot is trying to reach a goal. At each timestep the robot observes image $y$ where the goal position is marked by a white pixel; here we show an example of one of these images. The goal position and robot position are randomly sampled at the start of each new interaction. (Right) Average distance between the goal and the robot's final position over $25$ trials. Shaded regions show SEM.}
		\label{fig:sim3}
	\end{center}
    \vspace{-1.5em}
\end{figure}

Our final simulation examines whether Stable-BC extends to visual settings.
Here a point mass robot attempts to reach a $2$D goal location. 
The robot's state $x \in \mathbb{R}^2$ is its position, and $x$ updates with linear dynamics $\dot{x} = u$.
In each interaction the start and goal position are uniformly randomly sampled. 
However, the robot is not given direct access to the goal; instead, the robot observes an image $y$ that displays the goal location in pixel space (see \fig{sim3}).
The robot must learn a policy that moves towards this goal based on its current position $x$ and the visual observation $y$. 

\p{Methods} The images that the robot receives have $21 \times 21$ pixels.
Offline, a simulated expert shows the robot what actions to take in respond to these images: each demonstration consists of a ($x$, $y$, $u$) pair (i.e., at state $x$, if the robot observes image $y$, it should take action $u$).
After collecting the set of these demonstrations, we first train an autoencoder that embeds images $y$ into a $10$-dimensional latent space using the encoder $\mathcal{E}(y)$.
We then apply \textbf{BC} and \textbf{Stable-BC} to learn policies of the form $\pi(x, \mathcal{E}(y))$.
Both methods are trained using the same demonstration data.

\p{Results} 
In \fig{sim3} we plot the distance between the robot's position at the end of each interaction and the goal state (i.e., Final State Error).
If the robot moves completely randomly, the expected Final State Error is $10$ units.
Our results from this proof-of-concept simulation suggest that Stable-BC can be applied to settings where $y$ consists of visual observations; we find that Stable-BC outperforms standard BC when given the same amount of training data.
\vspace{-0.5em}
\section{Air Hockey Experiment} \label{sec:experiments}

\begin{figure*}
    \centering
    \includegraphics[width=1.9\columnwidth]{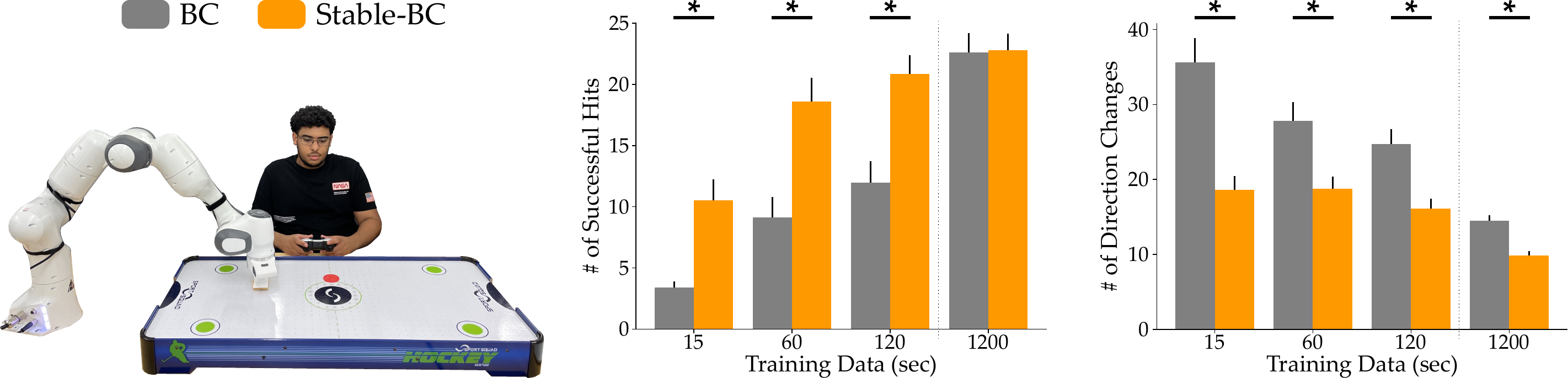}
    \vspace{-0.5em}
    \caption{Results for the air hockey experiment in Section \ref{sec:experiments}. (Left) Participants teleoperated a $7$ DoF robot arm to hit the puck. We collected their demonstration data offline, and then used this data to train BC and Stable-BC policies. (Center) We measured the number of successful hits with different amounts of training data. Ideally, a robust robot policy will repeatedly hit the puck, even when that puck travels with previously unseen angles and velocities. Both BC and Stable-BC eventually converged to equivalent performance, but Stable-BC reached that performance with a smaller amount of training data. (Right) To qualitatively assess the learned behavior, we also measured the number of direction changes per successful hit. Stable-BC produced policies that were more smooth and consistent, with fewer direction changes than BC. Error bars show SEM and $*$ denotes statistical significance ($p < 0.05$).}
    \vspace{-1.5em}
    \label{fig:experiments}
\end{figure*}

In this section we evaluate Stable-BC in a real-world environment with user-provided training data.
Specifically, we conduct imitation learning experiments where participants teach a 7-DoF Franka Emika robot arm to play a simplified game of air hockey.
We compare Algorithm~\ref{alg:M1} (Stable-BC) to standard behavior cloning (BC).
Videos of our air hockey experiments are available here: \href{https://youtu.be/ZC3BjY1k18w}{https://youtu.be/ZC3BjY1k18w}

\p{Experimental Setup}
The robot's task is to hit the hockey puck so that it bounces off the opposite side of the table and returns to the robot (see Figures~\ref{fig:front} and \ref{fig:experiments}).
The robot's state $x \in \mathbb{R}^2$ is the position of its end-effector on the surface of the air hockey table, and action $u \in \mathbb{R}^2$ is the robot's end-effector velocity.
A camera is mounted directly above the table to track the position of the hockey puck at a frame rate of $20$ Hz.
The environment state $y \in \mathbb{R}^4$ is the current and previous position of the puck in this camera frame; $y$ evolves with unknown dynamics $g(x, y, u)$.
Because the robot does not have access to $g$, in this experiment we applied the model-free version of our proposed Stable-BC algorithm.

\p{Training Data}
We recruited $10$ members of the Virginia Tech community to provide offline training data. Participants gave their informed consent under IRB $\#23$-$784$.
We first gave the participants $2$ minutes to practice controlling the robot and hitting the puck.
Once this practice was complete, each participant teleoperated the robot to repeatedly hit the puck against the opposite side of the table for $\sim 2.5$ minutes.
This resulted in $\sim 3000$ state-action pairs per user.
We kept each user's data separate, so that we obtained $10$ different datasets $\mathcal{D}_1 \ldots \mathcal{D}_{10}$ that we used to test our approach.

\p{Testing Procedure}
Given the expert datasets $\mathcal{D}_1 \ldots \mathcal{D}_{10}$, we trained robot policies using BC and Stable-BC.
We varied the amount of data the robot had access to during training --- e.g., we trained robot policies with $15$, $60$, and $120$ seconds of expert data.
For each amount of training data we learned $10$ different policies (one for every user's dataset), and then we tested the performance of each policy across $10$ independent rollouts.
The proctor started every trial by pushing the puck towards the robot, and then the robot executed its policy to autonomously and repeatedly hit the puck.

We quantified the performance of the robot learner by measuring the average number of times that the robot consecutively hit the puck against the opposite side of the table without missing it (\textit{Number of Successful Hits}).
If the robot successfully hit the puck $25$ times in a row, we terminated the trial there; i.e., $25$ was the maximum possible number of successful hits.
To assess the quality of the robot's motion, we also measured the \textit{Number of Direction Changes} per successful hit.
A direction change was defined as a difference between actions $u^t$ and $u^{t-1}$ of more than $10$ degrees.

\p{Hypothesis}
We had the following two hypotheses:\\
\textbf{H1.} \textit{Given the same training data, Stable-BC will achieve a higher number of successful hits as compared to BC.}\\
\textbf{H2.} \textit{Stable-BC will learn policies that output actions with fewer direction changes.}

\p{Results}
Our results are summarized in \fig{experiments}. 
A repeated measures ANOVA revealed that both the robot's learning algorithm ($F(1, 9) = 19.03$, $p<0.05$) and the amount of training data ($F(2, 18) = 50.79$, $p<0.05$) had significant effects on the number of successful hits.
For $15$, $60$, and $120$ seconds of training data, Stable-BC resulted in more robust policies that had a higher number of successful hits than BC ($p < 0.05$). 
As expected, the performance of both imitation learning algorithms increased in proportion to the amount of training data.
But Stable-BC was able to converge to ideal performance with less data than BC: under Stable-BC, the number of successful hits with $120$ seconds ($2$ minutes) of data was only marginally less than the number of successful hits with $1200$ seconds ($20$ minutes) of training data.
Both BC and Stable-BC converged to similar performance when given $1200$ seconds of data --- i.e., the combined data across all $10$ users --- indicating that Stable-BC is as effective or more effective than BC across all data levels.
Overall, these results support hypothesis \textbf{H1}.

We next explored the smoothness of the robot's learned policy.
As before, a repeated measures ANOVA found that the robot's learning algorithm ($F(1, 9) = 59.65$, $p<0.05$) as well as the amount of training data  ($F(2, 18) = 9.62$, $p<0.05$) had significant effects of the number of direction changes. 
Post-hoc analysis confirmed that across all levels of learning data, Stable-BC produced policies with significantly fewer direction changes ($p<0.05$) than BC.
We even observed that Stable-BC had fewer direction change when trained on the combined dataset with $1200$ seconds of data ($t(9) = 4.681, p < 0.05$).
Viewed together, these results support hypothesis \textbf{H2} and suggest that not only does Stable-BC lead to more robust policies, but these policies are qualitatively more smooth and consistent.

\vspace{-0.5em}
\section{Conclusion}

In this paper we presented a behavior cloning approach to reduce covariate shift.
Instead of focusing on the training data, our method explored on the error dynamics between the robot's current behavior and the expert's demonstrated behaviors.
By performing control theoretic analysis on these dynamics, we derived model-based and model-free stability conditions for shaping the learned policy to bound covariate shift.
Our resulting algorithm, Stable-BC, is an easy to implement extension of standard behavior cloning that can be used independently or alongside existing data-centric approaches.
Multiple experiments across interactive, nonlinear, visual, and real-world environments suggest that Stable-BC produces more robust policies than state-of-the-art baselines given the same training data.

\vspace{-0.5em}
\bibliography{references} 
\bibliographystyle{ieeetr}
\clearpage
\end{document}